\documentclass{article}
\usepackage{spconf,amsmath,graphicx,amssymb}

\title{Feature selection via simultaneous sparse approximation for person specific face verification}
\name{Yixiong~Liang,~Lei~Wang,~Shenghui~Liao,~Beiji~Zou}
\address{School of Information Science and Engineering, Central
South University, \\Changsha, Hunan 410083, China\\ \{yxliang,
wanglei, lsh, bjzou\}@mail.csu.edu.cn}
\begin{document}
%
\maketitle
\begin{abstract}
There is an increasing use of some imperceivable and redundant local
features for face recognition. While only a relatively small
fraction of them is relevant to the final recognition task, the
feature selection is a crucial and necessary step to select the most
discriminant ones to obtain a compact face representation. In this
paper, we investigate the sparsity-enforced regularization-based
feature selection methods and propose a multi-task feature selection
method for building person specific models for face verification. We
assume that the person specific models share a common subset of
features and novelly reformulated the common subset selection
problem as a simultaneous sparse approximation problem. The
effectiveness of the proposed methods is verified with the
challenging LFW face databases.
\end{abstract}
\begin{keywords}
Person specific face verification, feature selection, multi-task
learning, simultaneous sparse approximation
\end{keywords}
\section{Introduction}
\label{sec:intro}


Although face recognition has achieved significant progress under
controlled conditions in the past decades, it is still a very
challenging problem in the uncontrolled environment such as the web
where pose, lighting, expression, age, occlusion and makeup
variations are more complicated. As local areas are often more
descriptive and more appropriate for dealing with those variations,
there is an increasing use of some imperceivable local features for
face verification. Those local descriptors are generally extracted
by performing some transformation (both linear or nonlinear) on the
local region only or followed by some explicitly spatial pooling
means such as the spatial histogramming scheme \cite{Brown2010}.
These initial representation is often redundant or over-completed,
whereas only a relatively small fraction of them is relevant to the
recognition task. Thus feature selection is a crucial and necessary
step to select the most discriminant ones from the local features to
obtain a compact face representation, which can not only improve
performance but also decrease the computational burden.

Adaboost-based method is the most popular and impressive feature
selection methods in face recognition Scenario
\cite{Jones2003,Zhang2004,Yang2004,Wang2009}. It applies the simple
weak classifier, which only consists in a threshold on the value of
a single feature, many times on differently weighted version of data
and therefore obtaining a sequence of weak classifiers corresponding
to the selected features. One possible problem of these methods is
very time consuming because of the need of training and evaluating a
different classifier for each feature. An alternative is the
sparsity-enforced regularization techniques \cite{Hastie2009} which
is the state-of-the-art feature selection tool in bioinformatics and
recently has been successful applied in face detection and
verification \cite{Destrero2009a}. The main advantages of the
regularization approach are its effectiveness even in the high
dimensionality small sample size cases coupled with the support of
well-grounded theory \cite{Hastie2009}.

The concern of this work is mainly about how to build person
specific models for both feature selection and face verification in
unconstrained environments. In this case, although the face
verification can be seen as a binary classification problem (accept
or reject), it is in fact several binary classification problems
(one for each client model) and thus its essence is by nature a
multiple binary classification problem. Most existing approaches
train a generic model for all individuals
\cite{Jones2003,Zhang2004,Yang2004}, which may fail to capture the
variations among different individuals and therefore are suboptimal.
Other approaches build person specific models for different
individuals separately \cite{Destrero2009a} and often lead to
overfitting due to the small sample size of each individual. To
combat over the overfitting problem, Wang et al. \cite{Wang2009}
adopted multi-task learning to improve the generalization
performance of the Adaboost-based methods.

In this paper, we investigate the multi-task generalization of
regularized methods and propose a multi-task feature selection
method for person specific face verification. We assume that
different person specific models share a common subset of relevant
features and novelly reformulate the common subset selection problem
as a simultaneous sparse approximation problem. The classification
can be done by simple linear regression such as ridge regression.
The experiment results on the LFW face database \cite{Huang2007}
demonstrate the advantages and effectiveness of the proposed
methods.

\section{Notation and setup}
\label{sec:NS}

Suppose that there are $L$ individuals to be verified. Given a
training image set of size $N$, among them $N_{l}$ images correspond
to the subject $l$, while the remaining images are of other subjects
excluding the known $L$ subjects. From each image we can obtain a
$d$-dimensional feature vector $\mathbf{f}$. Let
$\mathbf{X}=[\mathbf{x}_{1},\dots,\mathbf{x}_{d}]\in
\mathbb{R}^{N\times d}$ be the data matrix with each row an input
feature vector, and
$\mathbf{Y}=[\mathbf{y}_{1},\dots,\mathbf{y}_{L}]\in
\mathbb{R}^{N\times L}$ be the corresponding indicator response
matrix where $\mathbf{y}_{l}$ is a $N$-dimensional vector with its
$i$th entry equal to 1 if the $i$th samples come from the subject
$l$ and else equal to 0. Therefore $\mathbf{Y}$ is a matrix of $0's$
and $1's$ with each row having at most a single $1$. We write
$\mathbf{c}_{l}$ for the $l$-th column of the matrix $\mathbf{C}$
and $\mathbf{c}^{l}$ for the $l$-th row.

\section{Single-task feature selection}
\label{sec:STFS}

We restrict ourselves to the case of regression models that are
linear in the components of feature. For class $l$, this linear
relationship can be characterized in matrix notation

\begin{equation}\label{eq1}
    \mathbf{y}_{l}=\mathbf{X}\mathbf{c}_{l}+b_{l}\mathbf{1},(l=1,\dots,L)
\end{equation}
where $\mathbf{c}_{l}$ is a $d$-dimensional coefficient vector and
$b_{l}$ is the \emph{bias} in the model of class $l$ respectively,
while $\mathbf{1}$ being a vector with its entries equal to 1. The
square error loss function
\begin{equation}\label{eq2}
    \textrm{Err}(\mathbf{c}_{l},b_{l})=\|\mathbf{y}_{l}-\mathbf{X}\mathbf{c}_{l}-b_{l}\mathbf{1}\|_{2}^{2}
\end{equation}
is adopted to fit the above linear model to the given training set.
Minimizing the square error loss function directly yields a unique
solution known as the least squares solution, which is typically
non-sparse and thus do not provide the feature selection in the
sense. A natural generalization for feature selection is $l_{0}$
regularization
\begin{equation}\label{eq3}
    \min_{\mathbf{c}_{l},b_{l}}\textrm{Err}(\mathbf{c}_{l},b_{l})+\lambda\|\mathbf{c}_{l}\|_{0},
\end{equation}
where $\|\cdot\|_{0}$ is the $l_{0}$ \emph{quasi-norm} counting the
nonzero entries of a vector and $\lambda$ quantifies how much
improvement in the approximation error is necessary before we admit
an additional term into the approximation. It is a classic
combinatorial sparse approximation problem which is a NP-hard in
general. A lot of numeric methods has been proposed to solve the
above combinatorial sparse approximation problem and two most common
approaches are greedy methods and convex relaxation methods. Greedy
techniques such as OMP abandon exhaustive search but iteratively
construct a sparse approximate one step at a time by selecting the
columns maximally reduces the residual and use it to update the
current approximation. Convex relaxation methods replace the
combinatorial sparse approximation problems with a related convex
version that can be solved more efficiently. As the $l_{1}$
\emph{norm} provides a natural convex relaxation of the $l_{0}$
\emph{quasi-norm}, the basis pursuit (BP) method solves the sparse
approximation problem by introducing an $l_{1}$ \emph{norm} in place
of the $l_{0}$ \emph{quasi-norm}
\begin{equation}\label{eq4}
    \min_{\mathbf{c}_{l},b_{l}}\textrm{Err}(\mathbf{c}_{l},b_{l})+\lambda\|\mathbf{c}_{l}\|_{1},
\end{equation}
which is an unconstrained convex function and can be solved by some
standard mathematical programming softwares. Similarly, the
parameter $\lambda$ negotiates a compromise between approximation
error and sparsity. It is also known as LASSO \cite{Hastie2009}.

Provided the regularization coefficient $\lambda$ is same across
different individuals, then solving each of these problems
independently is equivalent to solving the global problem obtained
by summing the objectives:
\begin{equation}\label{eq5}
    \min_{\mathbf{C},\mathbf{b}}\sum_{l=1}^{L}\frac{1}{N_{l}}\textrm{Err}(\mathbf{c}_{l},b_{l})+\lambda\sum_{l=1}^{L}\|\mathbf{c}_{l}\|_{0},
\end{equation}
Where $\mathbf{C}$ is the coefficient matrix with $\mathbf{c}_{l}$
in columns and $\mathbf{b}=[b_{1},\dots,b_{L}]^{T}$ is the bias
vector. Similarly, the corresponding $l_1$ \emph{norm} relaxation
objective is
\begin{equation}\label{eq6}
    \min_{\mathbf{C},\mathbf{b}}\sum_{l=1}^{L}\frac{1}{N_{l}}\textrm{Err}(\mathbf{c}_{l},b_{l})+\lambda\sum_{l=1}^{L}\|\mathbf{c}_{l}\|_{1}.
\end{equation}
%

\section{Multi-task feature selection}
\label{sec:MTFS}

In this section, we will describe our proposed multi-task feature
selection in details. As mentioned before, we assume each face
shares common subset of the redundant and imperceivable local
features. It's reasonable because each face shares a common
structure, i.e. face is composed of eyebrow, eye, nose, mouth, etc.
In the regularized feature selection frame, sharing a small subset
of features means that the coefficient matrix $\mathbf{C}$ has many
rows which are identically equal to zero and the corresponding
features will not be used for all tasks. Thus the global common
feature selection can be formulated as searching minimum number of
nonzero rows of $\mathbf{C}$ while balancing the error loss function
\begin{equation}\label{eq7}
    \min_{\mathbf{C},\mathbf{b}}\sum_{l=1}^{L}\frac{1}{N_{l}}\textrm{Err}(\mathbf{c}_{l},b_{l})+\lambda\|\mathbf{C}\|_{row-l_{0}},
\end{equation}
where $\|\cdot\|_{row-l_{0}}$ is the row-$l_{0}$ \emph{quasi-norm}
which denotes the number of nonzero rows and is given by
\begin{equation}\label{eq8}
\|\mathbf{C}\|_{row-l_{0}}=|\bigcup_{l=1}^{L}\mathrm{supp}(\mathbf{c}_{l})|,
\end{equation}
where $\mathrm{supp}(\cdot)$ denotes the support of a vector. When
the matrix $\mathbf{C}$ is a column vector, the row-support
degenerates to the support of the vector and the row-$l_{0}$
\emph{quasi-norm} degenerates to the usual $l_{0}$
\emph{quasi-norm}. If we regard $\mathbf{X}$ as a dictionary and
$\mathbf{y}_{l}(l=1,\cdots,L)$ as a serious of signals to be
approximated, the problem (\ref{eq7}) is indeed a simultaneous
sparse approximation problem.

It is immediately clear that the combinatorial optimization problem
(\ref{eq7}) is at least as hard as combinatorial optimization
problem (\ref{eq3}) and thus it is a more complicated NP-hard
problem in general. Some greedy pursuit algorithms such as
simultaneous orthogonal matching pursuit (SOMP) \cite{Tropp2006a}
are proposed to solve this combinatorial optimization problem.
Another approach to simultaneous sparse approximation is to replace
the row-$l_{0}$ \emph{quasi-norm} by a closely related convex
function \cite{Tropp2006b}. There are many different ways to relax
the row-$l_{0}$ \emph{quasi-norm} and one may define an entire
family of relaxations of the following form
\begin{equation}\label{eq9}
\|\mathbf{C}\|_{p,q}=\sum_{i=1}^{d}(\|\mathbf{c}^{l}\|_{q})^{p/q}=\sum_{i=1}^{d}[\sum_{j=1}^{L}|c_{ij}|^{q}]^{p/q}.
\end{equation}
This relaxation can be done by first applying the $l_{q}$
\emph{norm} to the rows of $\mathbf{C}$ and then applying the
$l_{p}$ \emph{norm} or \emph{quasi-norm} to the resulting vector of
$l_{p}$ \emph{norm}. On the one hand, we want to obtain row-sparse
of $\mathbf{C}$. On the other hand, we want the selected feature to
contribute to as many individuals as possible. This requires most
rows of $\mathbf{C}$ should be zero but the nonzero rows should have
many nonzero entries. Therefore we have $p\leq 1$ and $q> 1$. The
rational behind this is that minimizing the $l_{p}(p\le 1)$
\emph{norm} promotes sparsity whereas minimizing the $l_{q}(q>1)$
\emph{norm} promotes non-sparsity. If set $p=1$ and $q=2$, our
method is equivalent to the multi-task feature selection frame
proposed in \cite{Obozinski2009}. In our implementation, we set
$q=\infty$ since the $l_{\infty}$ \emph{norm} can provide better
non-sparsity than $l_{2}$ \emph{norm}. Replacing the row-$l_{0}$
\emph{quasi-norm} in the combinatorial optimization problem
(\ref{eq7}) by its relaxation (\ref{eq9}) with $p=1$ leads to the
following convex program
\begin{equation}\label{eq10}
    \min_{\mathbf{C},\mathbf{b}}\sum_{l=1}^{L}\frac{1}{N_{l}}\textrm{Err}(\mathbf{c}_{l},b_{l})+\lambda\|\mathbf{C}\|_{p,q},
\end{equation}
which can be solved by some standard mathematical programming
software \cite{Tropp2006b}.

Recalled that the above feature selection frame can be used for
classification directly in that it fits linear regression models to
the class indicator variables. One can also consider its usage as a
pure feature selection tool and explore some other common
classifiers for classification. Moreover, the proposed method is not
specific for face verification but to any other classification or
regression problem, providing that the tasks share the same training
data.

\section{Experimental results}
\label{sec:results}

We carry out some experiments on the LFW face database
\cite{Huang2007}. The LFW face database contains $13,233$ labeled
face images collected from news sites in the Internet. These images
belong to $5,749$ different individuals and have high variations in
position, pose, lighting, background, camera and quality, which make
it appropriate to evaluate face verification methods in realistic
and unconstrained environments. As there is not available protocol
along with the database for person specific face verification, we
select $158$ people with at least 10 images in the database as the
known people, i.e. $L=158$. For each known people, we choose the
former $5$ images for training and the remaining for testing. We
also select $210$ people with only one image in the database as the
background person (or unknown person) for training. Hence we have a
training set of size $1,000$ corresponding to $368$ people and a
testing set of size $3,534$ from the known $158$ people. Note that
our training set is overwhelmingly \emph{unbalanced} ($5$ positive
samples and $995$ negative samples with their ratio be close to
$1:200$ for each person).

In our experiments, each image is rotated and scaled so that centers
of the eyes are placed on specific pixels and then was cropped to
$64\times64$ pixels. We choose Gabor feature as the initial
representation due to its peculiar ability to model the spatial
summation properties of the receptive fields of the so called "bar
cells" in the primary visual cortex. We use 40 Gabor filters with
five scales $\{0,\cdots,4\}$ and eight orientations $\{0,\cdots,7\}$
which are common in face recognition area to obtain the Gabor
feature. The dimension $d$ of the resulting feature is then
$64\times64\times40 = 163,840$.

We apply both single-task and multi-task feature selection approach
to select the most informative $300$ features from the original
$163,840$-dimensional Gabor features. From a run-time point of view,
OMP and SOMP are adopted to solve the single-task and multi-task
feature selection problems, respectively. The outputs of OMP and
SOMP include both the indexes of the selected features and the
corresponding weights and therefore can be used for verification
directly. We also utilize the ridge regression method to determine
the weights of the selected features. The corresponding verification
methods are denoted as "STL", "MTL" and "STL+R" and "MTL+R". In
addition, we adopt the Adaboost-based method as the baseline for
feature selection and verification.

\begin{figure}[tb]

\begin{minipage}[b]{0.9\linewidth}
  \centering
  \centerline{\includegraphics[width=8cm]{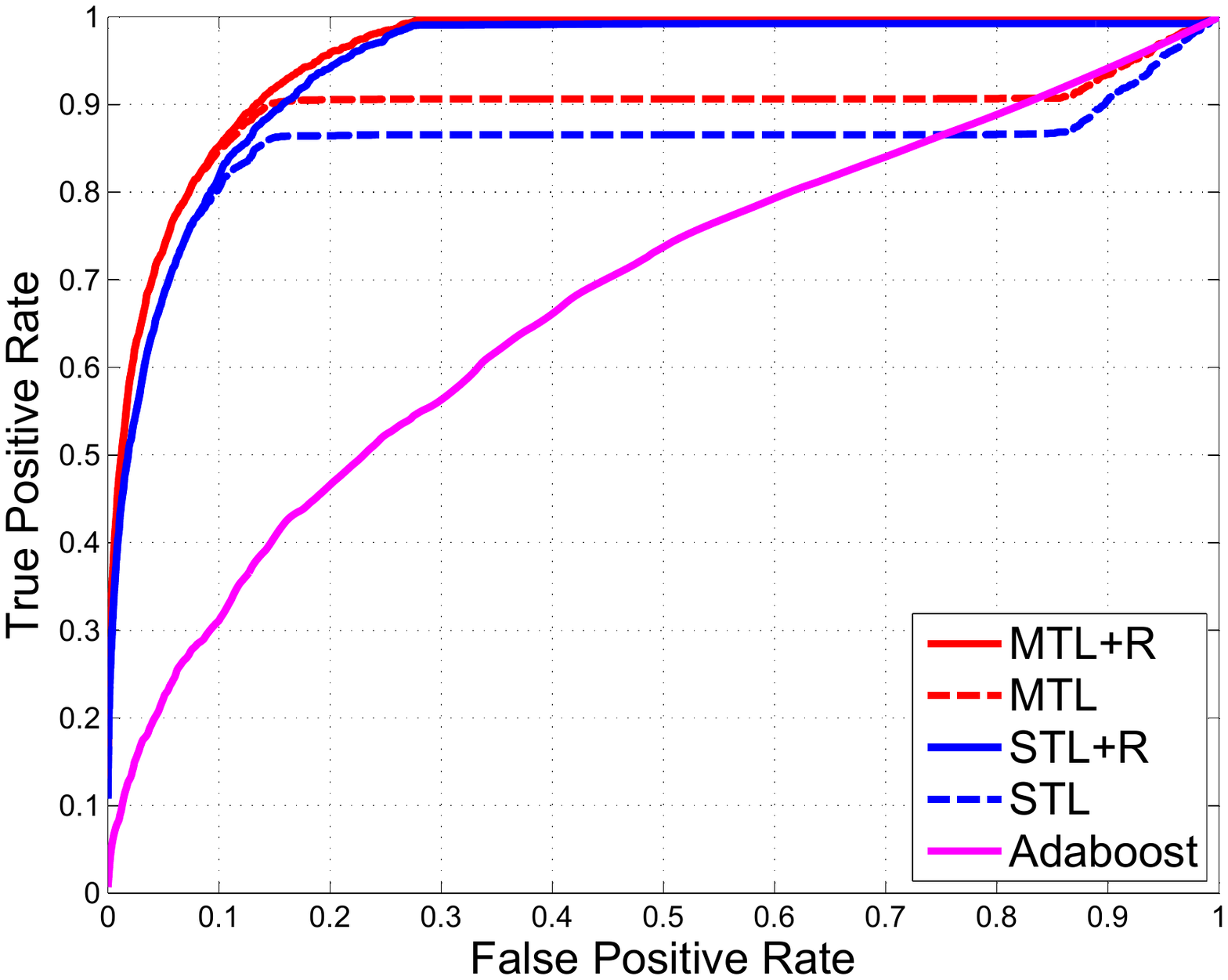}}
\end{minipage}
\caption{Average ROC curves of verifying images of 158 known people
only using $300$ Gabor features} \label{fig:AvgROC}
\end{figure}

Those methods all can verify the training set exactly, but perform
very differently on the testing set. We adopted the average ROC
curves and the average area under ROC curves (AUC) to evaluate their
performance across different persons. The comparative performance is
shown in Fig. \ref{fig:AvgROC} and Table \ref{tab:tab1}. The
Adaboost-based method may suffer from the unbalance of the training
set and performs much worse than the regularization-based methods.
The proposed multi-task feature selection methods ("MTL" and
"MTL+R") perform better than the corresponding single-task feature
selection methods ("STL" and "STL+R"). Another observation is that
the ridge regression-based verification does marginally improve the
performance compared with directly using the feature selection frame
for verification. This can be attributed to the fact that the
sparsity-enforcement in the feature selection frame may
underestimate the resulting coefficients and hereby obtain the worse
performance.

\begin{table}[t]
\caption{The average true positive rates (TPR) using different
methods when the false positive rate (FPR) is fixed at 0.1 and the
average AUC} \label{tab:tab1} \centering
\begin{tabular}{|l|c|c|}
  \hline
  Methods & TPR(std. dev.) & AUC(std. dev.)\\
    \hline
STL & 0.8046($\pm$ 0.1600) & 0.8506($\pm$ 0.1255) \\
MTL & 0.8465($\pm$ 0.1458) & 0.8901($\pm$ 0.0969) \\
STL+R & 0.8185($\pm$ 0.1636) & 0.9444($\pm$ 0.1458) \\
MTL+R & \textbf{0.8525($\pm$ 0.1480)} & \textbf{0.9586($\pm$ 0.0288)} \\
\hline
Adaboost & 0.3112($\pm$ 0.1708) & 0.6811($\pm$ 0.1066) \\
  \hline
\end{tabular}
\end{table}

\section{Conclusions}
We have proposed a multi-task learning method for building of
personal specific models both for feature selection and face
verification. The person specific models are jointly learned by
sharing the training data and then the multi-task feature selection
problem can be reformulated as a simultaneous sparse approximation
problem which can be solved by some greedy algorithms such as SOMP
or some related convex relaxation methods. The experimental results
show that the proposed multi-task feature selection method can
overcome the potential overfitting issues due to the lack of
training data and the adoption of ridge regression for verification
can marginally improve the performance.

\section{ACKNOWLEDGEMENT} \label{sec:ACK}
This research is partially supported by National Natural Science
Funds of China (60803024, 60970098 and 60903136), Specialized
Research Fund for the Doctoral Program of Higher Education
(200805331107 and 20090162110055), Fundamental Research Funds for
the Central Universities (201021200062), Open Project Program of the
State Key Lab of CAD\&CG, Zhejiang University (A0911 and A1011).

\bibliographystyle{IEEEbib}
\bibliography{strings,refs}

\begin{thebibliography}{10}

\bibitem{Brown2010}
M.~Brown, G.~Hua, and S.~Winder,
\newblock ``Discriminative learning of local image descriptors,''
\newblock {\em TPAMI}, vol. 33, pp. 43--57, 2011.

\bibitem{Jones2003}
M.~Jones and P.~Viola,
\newblock ``Face recognition using boosted local features,''
\newblock in {\em ICCV}. IEEE, 2003.

\bibitem{Zhang2004}
G.C. Zhang, X.S. Huang, S.Z. Li, Y.S.Wang, and X.H. Wu,
\newblock ``Boosting local binary pattern (lbp)-based face recognition,''
\newblock in {\em Proc. Advances in Biometric Person Authentication}, 2004, pp.
  179--186.

\bibitem{Yang2004}
P.~Yang, S.G. Shan, W.~Gao, S.Z. Li, and D.~Zhang,
\newblock ``Face recognition using ada-boosted gabor features,''
\newblock in {\em AFGR}. IEEE, 2004, pp. 356--361.

\bibitem{Wang2009}
X.G. Wang, C.~Zhang, and Z.Y. Zhang,
\newblock ``Boosted multi-task learning for face verification with applications
  to web images and video search,''
\newblock in {\em CVPR}. IEEE, 2009, pp. 142--149.

\bibitem{Hastie2009}
T.~Hastie, R.~Tibshirani, and J.~Friedman,
\newblock ``The elements of statistical learning: Data mining, inference, and
  prediction (2nd edition),''
\newblock {\em Springer}, 2009.

\bibitem{Destrero2009a}
A.~Destrero, C.De Mol, F.~Odone, and A.~Verri,
\newblock ``A regularized framework for feature selection in face detection and
  authentication,''
\newblock {\em Int. J. Comput. Vis.}, vol. 83, pp. 164--177, 2009.

\bibitem{Huang2007}
G.B. Huang, M.~Ramesh, T.~Berg, and E.~Learned-Miller,
\newblock ``Labeled faces in the wild: A database for studying face recognition
  in unconstrained environments,''
\newblock {\em University of Massachusetts, Amherst, Technical Report 07-49},
  2007.

\bibitem{Tropp2006a}
J.A. Tropp, A.C. Gilbert, and M.J. Strauss,
\newblock ``Algorithms for simultaneous sparse approximation. part i: Greedy
  pursuit,''
\newblock {\em Signal Processing}, vol. 86, pp. 572--588, 2006.

\bibitem{Tropp2006b}
J.A. Tropp, A.C. Gilbert, and M.J. Strauss,
\newblock ``Algorithms for simultaneous sparse approximation. part ii: Convex
  relaxation,''
\newblock {\em Signal Processing}, vol. 86, pp. 589--602, 2006.

\bibitem{Obozinski2009}
G.~Obozinski, B.~Taskar, and M.I. Jordan,
\newblock ``Joint covariate selection and joint subspace selection for multiple
  classification problems,''
\newblock {\em Journal of Statistics and Computing}, pp. 1--22, 2009.

\end{thebibliography}

\end{document}